\documentclass[letterpaper]{article} 
\usepackage[T1]{fontenc}
\usepackage{iclr2027_conference,times}

\usepackage{amsmath,amsfonts,bm}

\def\eqref#1{equation~\ref{#1}}

\def\1{\bm{1}}

\DeclareMathAlphabet{\mathsfit}{\encodingdefault}{\sfdefault}{m}{sl}
\SetMathAlphabet{\mathsfit}{bold}{\encodingdefault}{\sfdefault}{bx}{n}

\usepackage{amssymb}
\usepackage{amsthm}
\usepackage{booktabs}
\usepackage{graphicx}
\usepackage{flafter} 
\usepackage{placeins} 
\usepackage{enumitem}
\usepackage{xurl}
\usepackage{hyperref}
\hypersetup{hidelinks}
\usepackage{caption}
\usepackage{float}
\usepackage{algorithm}
\usepackage{algpseudocode}
\title{GPDPO: Learning from Reward Conflicts via Group Pareto-Dominance Policy Optimization}

\author{Zixuan Xu\textsuperscript{1}\thanks{Equal contribution.} \quad Haoxuan Chen\textsuperscript{1}\footnotemark[1] \quad Jiangxin Sun\textsuperscript{2} \quad Jian-Fang Hu\textsuperscript{1}\thanks{Corresponding author.} \\
\textsuperscript{1}ISEE Lab, Sun Yat-sen University \\
\textsuperscript{2}Computer Science and Technology, University of Trento \\
\texttt{\{chenhx253, xuzx57\}@mail2.sysu.edu.cn, hujf5@mail.sysu.edu.cn} \\
}

\iclrfinalcopy 
\begin{document}

\maketitle

\begin{abstract}
Multi-reward reinforcement learning offers a promising framework for aligning large language models (LLMs) with diverse objectives, but conflicting reward signals can induce ambiguous update directions and hinder stable optimization. Existing approaches address such conflicts through reward reweighting, gradient coordination, or rollout filtering, yet they primarily operate on per-response or group-level optimization signals and underexploit relational information among sampled responses. As a result, conflict-filtered rollouts may still contain useful supervision that is discarded during training.
We propose Group Pareto-Dominance Policy Optimization (GPDPO), a multi-reward policy optimization method that recovers pairwise supervision from conflict-filtered rollouts. GPDPO identifies Pareto-dominance relations among sampled responses and converts them into auxiliary preference signals, which are jointly optimized with the original policy objective. This enables more effective reuse of existing rollouts without additional sampling or reward evaluation.
Experiments on API-Bank across multiple model scales and reward configurations show that GPDPO consistently outperforms competitive multi-reward optimization baselines. Further analysis demonstrates that GPDPO reduces both group-relative and pairwise reward conflicts, supporting the effectiveness of exploiting relational supervision for multi-reward alignment. Code will be released.
\end{abstract}



\section{Introduction}
\label{sec:introduction}

Reinforcement learning provides an effective framework for improving large language models (LLMs) by optimizing task-specific rewards~\citep{shao2024deepseekmath,liu2026gdpo}. Group-based policy optimization methods such as GRPO~\citep{shao2024deepseekmath} construct policy updates from relative rewards among sampled responses. Extending this paradigm to multiple reward signals enables LLMs to jointly optimize diverse alignment objectives, such as task correctness, output format, and response length~\citep{liu2026gdpo,liu2026gd2po}. However, multi-reward optimization introduces a distinct challenge: different reward dimensions may provide conflicting signals for the same response.

Existing methods address such conflicts from several perspectives, including reward reweighting~\citep{yuan2026amrp}, gradient coordination~\citep{li2025gapo}, reward-wise advantage normalization~\citep{liu2026gdpo}, and conflict-aware rollout filtering~\citep{liu2026gd2po}. These approaches improve the compatibility of multiple reward signals by modifying how rewards, advantages, gradients, or samples contribute to policy optimization. In particular, conflict-aware filtering removes responses whose reward-wise advantages disagree in sign, preventing ambiguous signals from directly entering the policy update.
While effective for conflict removal, this response-level filtering is coarse: once a response is deemed conflicting, all supervision associated with it is discarded. This raises an important question: if a response is excluded from the policy update due to conflict-filtering, does that necessarily mean it contains no useful information for learning?

\begin{figure}[t]
  \centering
  \includegraphics[width=0.85\linewidth]{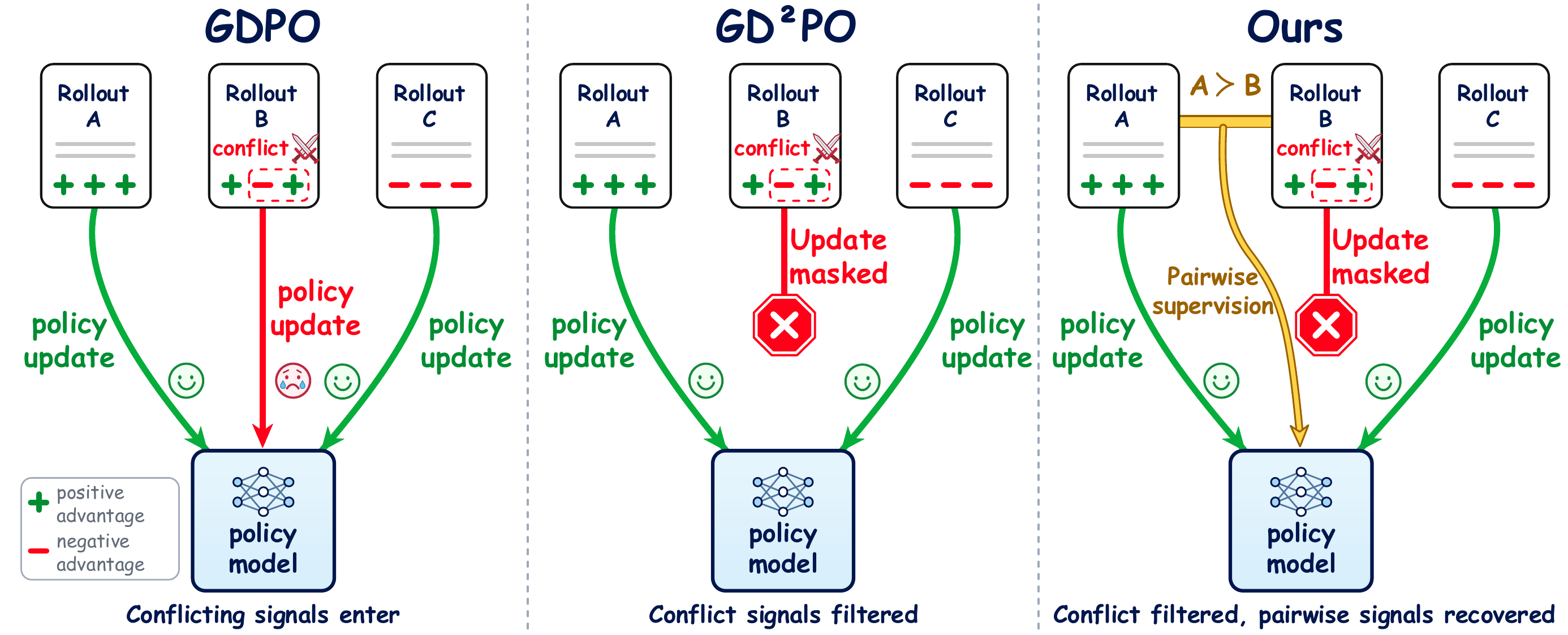}
  \caption{
  Different strategies for utilizing conflicting responses in multi-reward policy optimization. GDPO retains conflicting responses in policy updates, whereas GD$^2$PO masks them to avoid ambiguous optimization signals. GPDPO preserves conflict-aware masking while recovering pairwise supervision from Pareto-dominance relations, i.e., one response is no worse than another on all reward dimensions and strictly better on at least one.
  }
  \label{fig:intro}
\end{figure}

Our observation shows that it does not. A response with conflicting group-relative advantages may still form a reliable ordering relative to another. In particular, one response may consistently dominate another across all reward dimensions, yielding an unambiguous Pareto-consistent preference even when the conflicting response itself is unsuitable for a policy update. As illustrated in Figure~\ref{fig:intro}, this reveals a distinction between individual update and pairwise comparison: a response can be unreliable as a direct optimization signal while remaining informative as relational supervision.

Motivated by this, we introduce GPDPO, a multi-reward policy optimization framework that preserves conflict-aware masking for direct updates while recovering pairwise supervision from Pareto-dominance relations involving filtered rollouts. Unlike prior approaches that primarily modify per-response optimization signals, such as rewards, advantages, or gradients~\citep{yuan2026amrp,li2025gapo,liu2026gdpo,liu2026gd2po}, GPDPO explicitly exploits the relational structure among sampled responses. 
Specifically, GPDPO identifies Pareto-dominance relations among rollouts and transforms these relations involving filtered responses into a pairwise preference loss, which encourages the policy to favor responses with consistent improvements across reward dimensions. By leveraging such supervision from filtered rollouts, GPDPO extracts additional training signals from existing samples without extra sampling or reward evaluation.
This enables the policy to suppress ambiguous update directions while retaining informative pairwise preferences that would otherwise be discarded. Extensive experiments on API-Bank~\citep{li2023apibank}, across two Qwen3 model scales~\citep{yang2025qwen3} and two reward configurations, show that GPDPO consistently outperforms competitive baselines in multi-reward optimization.

Our contributions are summarized as follows: 1) We show that responses with conflicting group-relative advantages can still yield reliable pairwise preferences, revealing supervision overlooked by response-level conflict filtering. 2) We propose GPDPO, which preserves conflict-aware masking for direct updates while recovering pairwise supervision from filtered responses through Pareto-dominance relations. 3) Extensive experiments on API-Bank across varied model scales and reward configurations show that GPDPO consistently outperforms competitive multi-reward baselines.

\section{Related Work}
\label{sec:related-work}

\textbf{Reinforcement learning with verifiable rewards.}
RLVR has shown strong potential for improving LLM reasoning~\citep{shao2024deepseekmath,guo2025deepseek}. Prior work advances this paradigm through group-relative advantage estimation~\citep{shao2024deepseekmath}, sampling and objective refinement~\citep{yu2025dapo,liu2025understanding}, and sequence-level policy optimization~\citep{zheng2025gspo}. These methods improve the efficiency and stability of reasoning optimization. However, applying GRPO-style algorithms to multiple rewards through scalar aggregation may obscure reward-specific distinctions, limiting their effectiveness in multi-reward optimization~\citep{liu2026gdpo}.

\textbf{Multi-reward policy optimization.}
Prior work coordinates multiple rewards through fixed or adaptive weighting~\citep{zhou2024modpo,lu2026dynamic}, gradient balancing~\citep{li2025gapo}, and reward-wise or blockwise advantage estimation~\citep{liu2026gdpo,pavlenko2026blockwise}. GD$^2$PO further filters rollouts with conflicting reward-wise advantages~\citep{liu2026gd2po}. These methods mitigate multi-reward conflicts by adjusting how reward signals contribute to policy updates. However, response-level conflict handling may overlook consistent inter-response relations and discard useful supervision. GPDPO instead complements group-relative conflict handling with pairwise comparisons, recovering reward-consistent supervision from filtered rollouts.

\textbf{Pareto dominance relations.}
Pareto dominance is widely used in multi-objective search, including NSGA-II~\citep{emmerich2018tutorial,deb2002nsga2}. In LLM alignment, related methods use Pareto or reward-consistency criteria to assign trajectory advantages~\citep{li2026paretopo}, construct preference pairs~\citep{li2025sipo,xu2026rcs}, or downweight conflicting pairs~\citep{xie2026dpoc}. These methods use Pareto dominance to guide multi-objective learning. Insipred by these, GPDPO uses Pareto dominance to identify consistent comparison structures among rollouts and draws on these structures to provide supplementary supervision for multi-reward optimization.


\section{Method}
\label{sec:method}

In this section, we first introduce multi-reward optimization and the associated conflict-filtering mechanism (Section~\ref{sec:preliminary}). We then revisit the sign-based filtering criterion, identify two potential limitations, and empirically examine their effects (Section~\ref{sec:group-mean}). Building on these findings, we propose GPDPO, which exploits Pareto-dominance relations between responses to recover fine-grained pairwise supervision for improved conflict mitigation (Section~\ref{sec:lore}).

\subsection{Preliminaries}
\label{sec:preliminary}

\textbf{Multi-reward group-relative optimization.}
Given a prompt \(x\), the sampling policy \(\pi_{\theta_{\mathrm{old}}}\) generates a group of \(G\) responses, denoted by \(\mathcal G(x)=\{y_1,\ldots,y_G\}\). Each response \(y_n\) receives an \(M\)-dimensional reward vector \(\mathbf r_n=(r_n^1,\ldots,r_n^M)\), where \(r_n^i=r_i(x,y_n)\) and higher values are preferred on every reward dimension. A straightforward way to extend GRPO~\citep{shao2024deepseekmath} to the multi-reward setting is to aggregate the rewards before applying group-relative normalization:
\begin{equation}
R_n=\sum_{i=1}^{M}w_i r_n^i,
\qquad
\overline R=\frac{1}{G}\sum_{n=1}^{G}R_n,
\qquad
A_n^{\mathrm{GRPO}}=\frac{R_n-\overline R}{s_R}.
\label{eq:grpo-adv}
\end{equation}
Here, \(w_i\) is the weight of reward dimension \(i\), and \(s_R>0\) is the within-group normalization scale. The resulting scalar advantage is used in the clipped policy objective~\citep{schulman2017ppo} to update the probability of the corresponding response.
\begin{figure}[!t]
    \centering
    \includegraphics[width=0.98\linewidth]{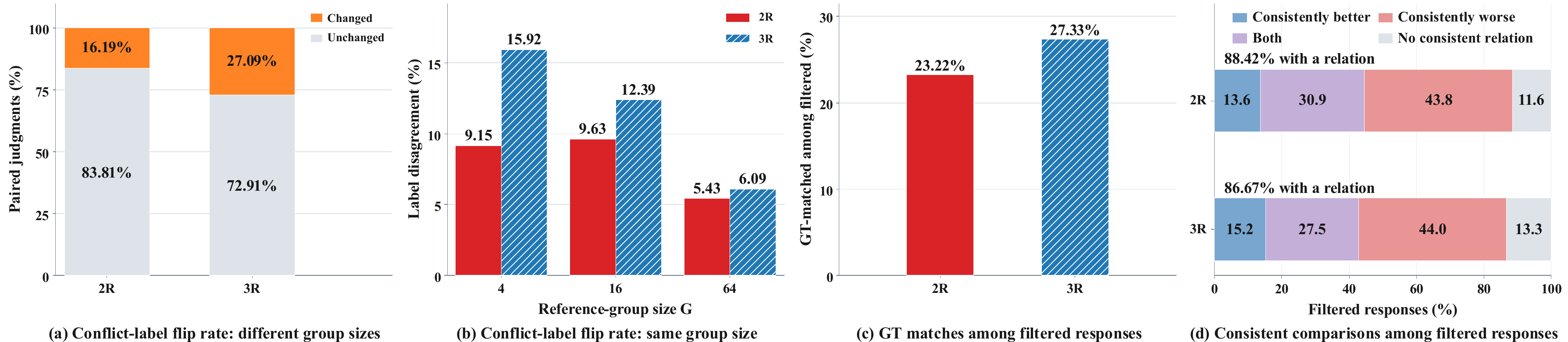}
    \caption{Conflict filtering is sensitive to the reference group and can discard useful learning signals. (a,b) Conflict-label flip rates for fixed responses under different group sizes and different peers at the same group size, respectively. (c) Percentage of filtered responses that match the ground-truth. (d) Consistent comparison relations between filtered responses and their peers. ``Both'' denotes being consistently better than some peers and consistently worse than others.}
    \label{fig:group-mean-study}
\end{figure}
GDPO~\citep{liu2026gdpo} observes that aggregating rewards too early can lose information specific to individual reward dimensions. It therefore normalizes each reward separately, aggregates the resulting advantages, and applies batch-level scale normalization. GD$^2$PO~\citep{liu2026gd2po} further addresses conflicts among these advantages: when a response has both positive and negative advantages across reward dimensions, direct aggregation can cause the learning signals to cancel each other out. GD$^2$PO therefore filters responses with conflicting advantage signs and adjusts the update through reweighting. Specifically, the reward-wise advantages and filtering gate are defined as
\begin{equation}
A_n^i=\frac{r_n^i-\mu_i}{s_i},
\qquad
\mu_i=\frac{1}{G}\sum_{j=1}^{G}r_j^i,
\qquad
g(y_n)=\mathbf 1\!\left\{
\operatorname{sign}(A_n^i)\operatorname{sign}(A_n^k)\geq0,\ \forall i,k
\right\}.
\label{eq:advantage-gate}
\end{equation}
Here, \(s_i\) denotes the normalization scale for reward dimension \(i\). The binary gate \(g(y_n)\) equals 1 for retained responses and 0 for filtered responses. Specifically, the gate selects responses whose reward-wise advantages share a consistent sign across dimensions, while removing the contribution of responses with conflicting directions. 


\textbf{Pareto dominance.}
Pareto dominance compares two vectors in a multi-objective space. \(\mathbf{a}\) Pareto-dominates \(\mathbf{b}\) if \(\mathbf{a}\) is at least as good as \(\mathbf{b}\) in all dimensions and strictly better in at least one dimension~\citep{emmerich2018tutorial}:
\begin{equation}
\mathbf a\succ_{\mathrm P}\mathbf b
\quad\Longleftrightarrow\quad
\bigl(\forall i:\ a_i\geq b_i\bigr)
\;\land\;
\bigl(\exists i:\ a_i>b_i\bigr).
\label{eq:pareto-definition}
\end{equation}
Pareto dominance does not allow gains on one objective to compensate for losses on another. Thus, when two vectors are better on different objectives, neither dominates the other.

\subsection{Rethinking Conflict Filtering in Multi-Reward RL}
\label{sec:group-mean}

Here, we revisit conflict filtering in multi-reward RL from the perspective of group-relative advantages. Our analysis reveals two limitations: filtering decisions are sensitive to the reference group, and responses removed due to group-wise conflicts may still contain useful pairwise supervision. Through empirical studies, we show that reward conflicts at the group level do not necessarily imply the absence of consistent response-level comparisons.


Prior work~\citep{liu2026gd2po} filters responses whose advantages disagree in sign across reward dimensions, which can be rewritten as follows:
\begin{equation}
\operatorname{sign}(A_n^i)
=\operatorname{sign}(r_n^i-\mu_i)
=\operatorname{sign}\!\left[
\sum_{j=1}^{G}(r_n^i-r_j^i)
\right].
\label{eq:atomic-decomposition}
\end{equation}
This formulation shows that the sign of a response's advantage depends on the sum of its reward differences from the other responses in the group. Conflict filtering based on these group-level signs has two limitations. First, \textbf{conflict filtering is sensitive to the reference group}: the same response can receive different filtering decisions as its peers change. Moreover, \textbf{group-wise conflict filtering can discard useful learning signals}. Coarse-grained conflict signals obtained by summing within-group reward differences cannot capture fine-grained comparisons between responses, leaving useful pairwise training signals unrecognized by the filter. To further analyze these limitations, we conduct the empirical experiments shown in Figure~\ref{fig:group-mean-study}. Detailed experiments are provided in the appendix.
\begin{figure}[!t]
    \centering
    \includegraphics[width=0.98\linewidth]{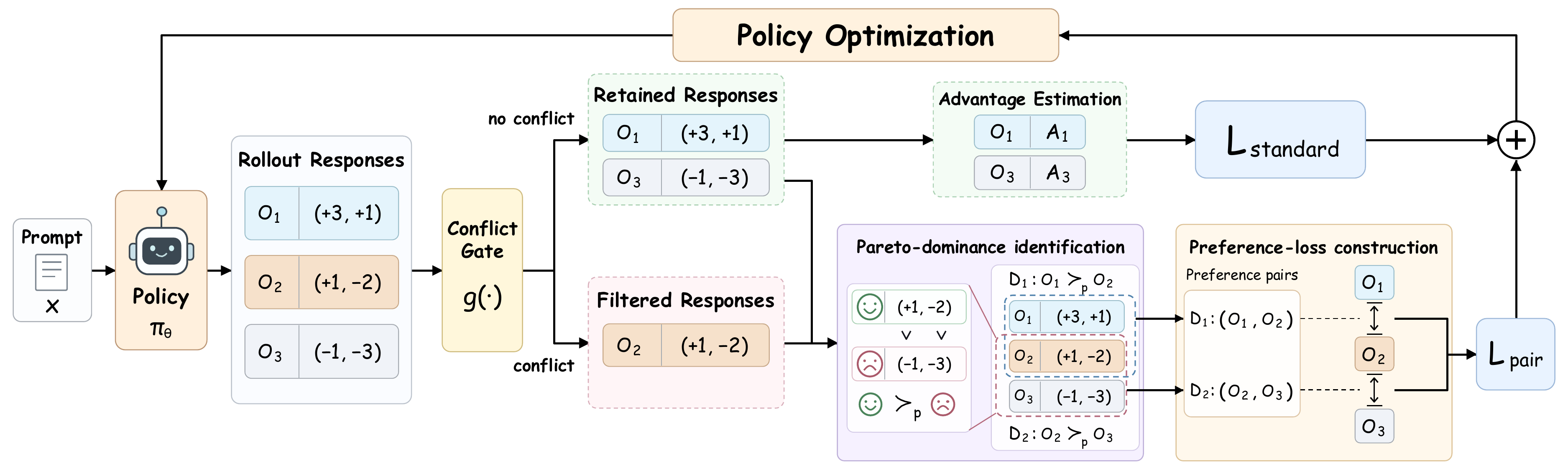}
    \caption{Overview of GPDPO. Based on conflict filtering, we identify Pareto dominance relations within filtered responses and leverage them to construct preference loss. The resulting preference loss is combined with the standard multi-reward policy-gradient objective for policy optimization.}
    \label{fig:method}
\end{figure}

\textbf{Conflict filtering is sensitive to the rollout group.}
We hold the target responses fixed, vary the reference group size, and examine whether the filtering decisions change. As shown in Figure~\ref{fig:group-mean-study}(a), expanding the original group from \(G=4\) to \(G=64\) changes the conflict filtering decisions for 16.19\% and 27.09\% of the fixed targets under 2R and 3R, respectively. Furthermore, we keep the group size fixed and replace the peer responses. Figure~\ref{fig:group-mean-study}(b) shows decision inconsistency rates of 9.15\% and 15.92\% at \(G=4\), which remain at 5.43\% and 6.09\% at \(G=64\). These results show that conflict filtering signals are highly sensitive to the reference group. 

\textbf{Group-wise conflict filtering can discard useful learning signals.}
We further analyze responses filtered due to reward conflicts. As shown in Figure~\ref{fig:group-mean-study}(c), 23.22\% and 27.33\% of these responses under 2R and 3R, respectively, exactly match the ground-truth (GT) ordered tool-call structure. This shows that filtered responses can still contain valuable training signals. We then investigate the fine-grained information in these responses. As shown in Figure~\ref{fig:group-mean-study}(d), 88.42\% and 86.67\% of filtered responses under 2R and 3R, respectively, have a consistent comparison relation with at least one other response in the group, yet these relations are discarded by group-wise conflict filtering. These findings show that group-wise conflict signals lose substantial fine-grained information, causing the filter to incorrectly discard many useful training samples and signals.

These observations lead to our motivation. First, conflict filtering signals should be decoupled from the group mean to improve training stability. Moreover, conflict detection should go beyond group-level relations to exploit comparisons between individual rollouts, which depend only on the properties of the two rollouts themselves.
Finally, even when a response has conflicting advantages relative to the group mean, pairwise comparisons with individual rollouts can still reveal consistent preference relations and provide useful supervision.

\subsection{GPDPO: Group Pareto-Dominance Policy Optimization}
\label{sec:lore}

In this section, we introduce GPDPO, a multi-reward policy optimization method that recovers pairwise supervision from conflict-filtered rollouts through Pareto dominance. Rather than relying solely on group-level conflict handling, GPDPO exploits consistent preference relations between individual responses and incorporates the resulting pairwise supervision into the original policy objective. As illustrated in Figure~\ref{fig:method}, GPDPO consists of three components: {(1) Pareto-dominance identification}, {(2) preference-loss construction}, and {(3) joint policy optimization}.

\textbf{Pareto-dominance identification.}
\label{sec:pareto}
Based on the conflict filtering mechanism of prior work~\citep{liu2026gd2po}, we obtain the filtered set \(\mathcal F(x)=\{y_n\in\mathcal G(x):c(y_n)=1\}\) and retained set \(\mathcal R(x)=\mathcal G(x)\setminus\mathcal F(x)\) from the sampled group \(\mathcal G(x)\). Here, \(c(y_n)=1-g(y_n)=1\) indicates that the response is filtered due to conflict. We define a consistent preference relation based on Pareto dominance~\citep{emmerich2018tutorial}. Following Section~\ref{sec:preliminary}, for two responses \(y_a\) and \(y_b\) to the same prompt \(x\), we say that \(y_a\) Pareto-dominates \(y_b\), if and only if \(\mathbf r_a \succ_{\mathrm P} \mathbf r_b\). For notational simplicity, we use \(y_a \succ_{\mathrm P} y_b\) to denote this relation between responses \(y_a\) and \(y_b\).

Pareto dominance depends only on the relative reward vectors of the two responses, independent of the group mean or other rollouts. This pairwise criterion therefore exposes reward-consistent, conflict-free comparison structures that can remain hidden under group-level aggregation. 




We collect all Pareto dominance relations within the sampled group into the following set:
\begin{equation}
\mathcal D(x)=\bigl\{(y_w,y_l)\in\mathcal G(x)^2:
y_w\succ_{\mathrm P}y_l\bigr\}.
\label{eq:dominance-pool}
\end{equation}
Here, \(y_w\) represents the dominating response and \(y_l\) represents the dominated response.

\textbf{Preference-loss construction.}
To convert the dominance set into optimization signals, we construct a preference-pair set by selecting dominance edges in \(\mathcal D(x)\) involving at least one response in the filtered set \(\mathcal F(x)\):
\begin{equation}
\mathcal E(x)=\bigl\{(y_w,y_l)\in\mathcal D(x):
y_w\in\mathcal F(x)\ \lor\ y_l\in\mathcal F(x)\bigr\}.
\label{eq:edge-pool}
\end{equation}
This selection focuses the auxiliary objective on supervision recovered from filtered rollouts, while retaining the original policy optimization pathway for non-filtered responses.

We then define the mean token-level log-probability ratio to measure the policy update of a response relative to the sampling policy:
\begin{equation}
u_\theta(y\mid x)
=\frac{1}{|y|}\sum_{t=1}^{|y|}
\log\frac{\pi_\theta(y_t\mid x,y_{<t})}
{\pi_{\theta_{\mathrm{old}}}(y_t\mid x,y_{<t})}.
\label{eq:sequence-ratio}
\end{equation}
Here, \(|y|\) denotes the number of tokens in the response, and \(\pi_{\theta_{\mathrm{old}}}\) is the sampling policy used to generate the rollout group. The score \(u_\theta(y\mid x)\) captures the relative change in the response likelihood under the updated policy.

To ensure that the auxiliary objective updates only on filtered responses, we stop gradients through retained responses when they participate in selected preference pairs:
\begin{equation}
\widetilde u_\theta(y\mid x)=
\begin{cases}
u_\theta(y\mid x), & y\in\mathcal F(x),\\
\operatorname{sg}[u_\theta(y\mid x)], & y\notin\mathcal F(x),
\end{cases}
\label{eq:filtered-score}
\end{equation}

where \(\operatorname{sg}\) denotes the stop-gradient operator. Retained responses serve as fixed comparison anchors, whereas filtered responses receive the preference supervision. When both endpoints of a preference pair are filtered, both responses contribute gradients to the auxiliary update.

For each selected dominance relation, we optimize the following pairwise preference objective:
\begin{equation}
\ell_{\mathrm{pair}}(y_w,y_l\mid x)
=-\log\sigma\!\left(
\beta\left[
\widetilde u_\theta(y_w\mid x)
-\widetilde u_\theta(y_l\mid x)
\right]\right),
\label{eq:edge-loss}
\end{equation}
where \(y_w\) and \(y_l\) denote the dominating and dominated responses, respectively. This objective encourages the policy to increase the relative likelihood of responses consistent with Pareto dominance, thereby incorporating reward-consistent supervision into policy optimization. Here, \(\sigma(\cdot)\) is the sigmoid function and \(\beta>0\) controls the preference margin.

\textbf{Joint policy optimization.}
For each training batch \(\mathcal B\), we assign equal weight to all selected
preference pairs and define the auxiliary objective as:
\begin{equation}
\mathcal L_{\mathrm{pair}}
=\frac{
\sum_{x\in\mathcal B}
\sum_{(y_w,y_l)\in\mathcal E(x)}
\ell_{\mathrm{pair}}(y_w,y_l\mid x)
}{
\sum_{x\in\mathcal B}|\mathcal E(x)|
}.
\label{eq:pair-aggregate}
\end{equation}
We finally formulate the GPDPO objective by combining the original training loss with the auxiliary pairwise objective, weighted by \(\lambda\):
\begin{equation}
\mathcal L_{\mathrm{GPDPO}}
=\mathcal L_{\mathrm{standard}}
+\lambda\mathcal L_{\mathrm{pair}},
\label{eq:joint-loss}
\end{equation}
where \(\mathcal L_{\mathrm{standard}}\) denotes the standard multi-reward policy-gradient objective used in GDPO~\citep{liu2026gdpo} and GD$^2$PO~\citep{liu2026gd2po}, and \(\lambda\) controls the contribution of the pairwise loss.

Our GPDPO recovers conflict-related supervision at a finer granularity by operating on pairwise relations between rollouts. Responses filtered by group-level conflict criterion may still form reward-consistent comparisons with specific peers. GPDPO identifies and exploits these fine-grained relations, reusing supervision already present in the rollout group without additional sampling or reward evaluation. The complete algorithm is provided in the appendix.

\section{Experiments}
\label{sec:experiments}

\subsection{Experimental Settings}
\label{sec:exp-settings}
\begin{table}[!t]
    \caption{Results on API-Bank under the two-reward (2R)  setting. Accuracy metrics are reported as percentages. The best and second-best results are highlighted in bold and \underline{underlined}, respectively.}
    \label{tab:main-2r}
    \centering
    \small
    \setlength{\tabcolsep}{3pt}
    \begin{tabular*}{\linewidth}{@{\extracolsep{\fill}}lrrrrrr@{}}
        \toprule
        Method & \shortstack{Level 1\\Acc.} & \shortstack{Level 2\\Acc.} & \shortstack{Level 3\\Acc.} & \shortstack{Correct\\Acc.} & \shortstack{Length\\Rew.} & Overall \\
        \midrule
\multicolumn{7}{l}{\textbf{Qwen3-1.7B}} \\
GRPO\citep{shao2024deepseekmath} & 66.17 & \underline{53.73} & \underline{46.56} & 60.47 & 0.9643 & 1.569 \\
GDPO~\citep{liu2026gdpo} & 66.67 & 61.19 & 43.51 & 60.97 & \textbf{0.9875} & 1.597 \\
GD$^2$PO~\citep{liu2026gd2po} & \underline{69.17} & 61.19 & 42.75 & \underline{62.48} & 0.9793 & \underline{1.604} \\
GPDPO & \textbf{70.93} & \textbf{61.19} & \textbf{48.85} & \textbf{64.99} & \underline{0.9812} & \textbf{1.631} \\
\midrule
\multicolumn{7}{l}{\textbf{Qwen3-4B}} \\
GRPO~\citep{shao2024deepseekmath} & \underline{71.93} & \underline{59.70} & 45.80 & 64.82 & 0.9898 & 1.638 \\
GDPO~\citep{liu2026gdpo} & \textbf{72.93} & 56.72 & \textbf{47.33} & \underline{65.49} & 0.9978 & \underline{1.653} \\
GD$^2$PO~\citep{liu2026gd2po} & 71.68 & \underline{59.70} & 45.80 & 64.66 & \underline{0.9987} & 1.645 \\
GPDPO & \textbf{72.93} & \textbf{64.18} & \underline{45.80} & \textbf{66.00} & \textbf{0.9995} & \textbf{1.659} \\
        \bottomrule
    \end{tabular*}
\end{table}

\begin{table}[!t]
    \caption{
Results on API-Bank under the three-reward (3R) setting. Accuracy metrics are reported as percentages. The best and second-best results are highlighted in bold and \underline{underlined}, respectively.
    }
    \label{tab:main-3r}
    \centering
    \small
    \setlength{\tabcolsep}{3pt}
    \begin{tabular*}{\linewidth}{@{\extracolsep{\fill}}lrrrrrrr@{}}
        \toprule
        Method & \shortstack{Level 1\\Acc.} & \shortstack{Level 2\\Acc.} & \shortstack{Level 3\\Acc.} & \shortstack{Correct\\Acc.} & \shortstack{Format\\Acc.} & \shortstack{Length\\Rew.} & Overall \\
        \midrule
\multicolumn{8}{l}{\textbf{Qwen3-1.7B}} \\
GRPO\citep{shao2024deepseekmath} & 65.91 & \textbf{62.69} & 39.69 & 59.80 & 98.99 & 0.9519 & 2.540 \\
GDPO~\citep{liu2026gdpo} & 68.92 & 53.73 & 41.22 & 61.14 & \textbf{100.00} & 0.9608 & 2.572 \\
GD$^2$PO~\citep{liu2026gd2po} & \underline{68.92} & 59.70 & \textbf{44.27} & \underline{62.48} & \textbf{100.00} & \underline{0.9756} & \underline{2.600} \\
GPDPO & \textbf{70.43} & \underline{61.19} & \underline{42.75} & \textbf{63.32} & \underline{99.66} & \textbf{0.9760} & \textbf{2.606} \\
\midrule
\multicolumn{8}{l}{\textbf{Qwen3-4B}} \\
GRPO\citep{shao2024deepseekmath} & 70.93 & 58.21 & \underline{46.56} & \underline{64.15} & 98.99 & 0.9874 & 2.619 \\
GDPO~\citep{liu2026gdpo} & 68.42 & \underline{56.72} & \underline{46.56} & 62.31 & \underline{99.83} & \textbf{0.9976} & 2.619 \\
GD$^2$PO~\citep{liu2026gd2po} & \underline{71.18} & \underline{56.72} & 42.75 & 63.32 & \textbf{100.00} & \underline{0.9967} & \underline{2.630} \\
GPDPO & \textbf{71.43} & \textbf{64.18} & \textbf{47.33} & \textbf{65.33} & \textbf{100.00} & 0.9839 & \textbf{2.637} \\
        \bottomrule
    \end{tabular*}
\end{table}
\paragraph{Task and rewards.}
We study multi-reward tool-call policy optimization on the 3,920 training examples from RLLA-4K~\citep{qian2025toolrl}. The upstream ToolRL training mixture draws on ToolACE~\citep{liu2025toolace}, Hammer (Masked)~\citep{lin2024hammer}, and xLAM~\citep{zhang2025xlam}. In the two-reward setting (2R), the reward consists of correctness and length; the three-reward setting (3R) additionally includes format. The correctness reward measures agreement between generated and annotated tool calls in tool names, argument names, and argument values, and is scaled to \([-3,3]\). The length reward is defined as \(\min(n_{\mathrm{think}}/512,1)\), where \(n_{\mathrm{think}}\) denotes the number of words inside the \texttt{<think>} block; responses without a \texttt{<think>} block receive zero. The format reward is binary and indicates whether the response follows the prescribed output-tag structure. Unless otherwise stated, all reward dimensions are weighted equally.

\textbf{Model training and evaluation.}
We evaluate Qwen3-1.7B and Qwen3-4B~\citep{yang2025qwen3}, comparing our GPDPO against GRPO~\citep{shao2024deepseekmath}, GDPO~\citep{liu2026gdpo}, and GD$^2$PO~\citep{liu2026gd2po}. GRPO aggregates reward dimensions before normalization, whereas GDPO normalizes each reward dimension separately before aggregation; GD$^2$PO further introduces conflict filtering and query-level reweighting. All methods are trained for 200 rollout steps with 512 prompts per step and \(G=4\) responses per prompt. We use one epoch per step, mini-batches of 128 prompts, and a learning rate of \(10^{-6}\). Responses are sampled with temperature 1.0, without a KL penalty or entropy bonus. The maximum prompt and response lengths are 2,048 and 1,024 tokens, respectively. Our implementation is built on verl~\citep{sheng2025hybridflow} and vLLM~\citep{kwon2023vllm}. Unless otherwise specified, GPDPO uses uniform pair weighting with \(\lambda=0.1\) and \(\beta=0.1\), and refreshes the sampling-policy reference after each rollout step.

We evaluate on API-Bank~\citep{li2023apibank}, which contains 597 questions across Levels 1/2/3, with 399/67/131 questions, respectively. Each row in the main results table corresponds to a single evaluation of the selected step-200 checkpoint using greedy decoding and a maximum generation length of 4,096 tokens. We report accuracy for each difficulty level, overall accuracy across all questions (\textsc{Correct Acc.}), format accuracy in the 3R setting, and mean length reward. Accuracy metrics are reported as percentages, while length reward remains on its original \([0,1]\) scale. The \textsc{Overall} score is defined as \(\textsc{Correct Acc.}/100 + \textsc{Length Rew.}\) for 2R, and additionally includes \(\textsc{Format Acc.}/100\) for 3R. We treat \textsc{Correct Acc.} as the primary measure of task performance.


\subsection{Main Results}
\label{sec:main-results}

\begin{figure}[!t]
    \centering
    \includegraphics[width=0.99\linewidth]{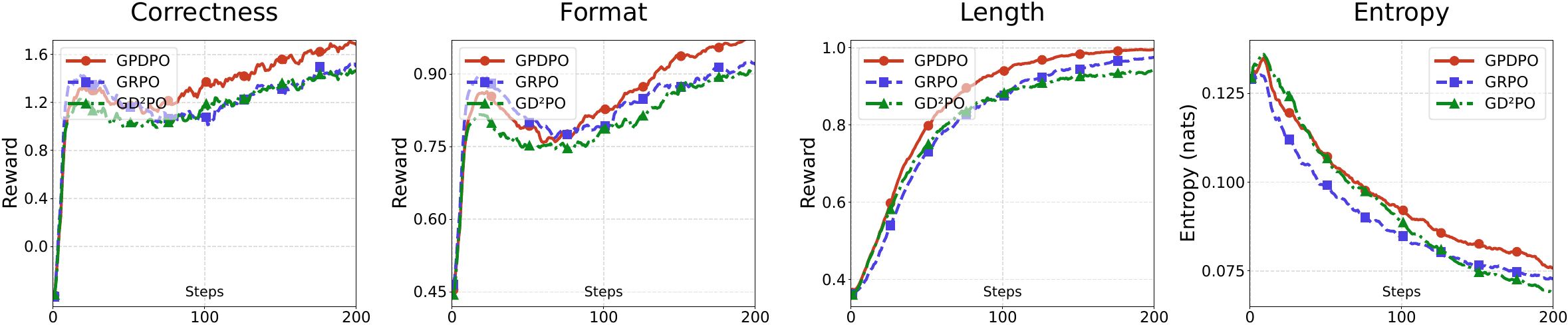}
    \caption{Training dynamics of Qwen3-1.7B under the three-reward (3R) setting.
    }
    \label{fig:training-dynamics}
\end{figure}

\begin{figure}[!t]
    \centering
 \includegraphics[width=0.89\linewidth]{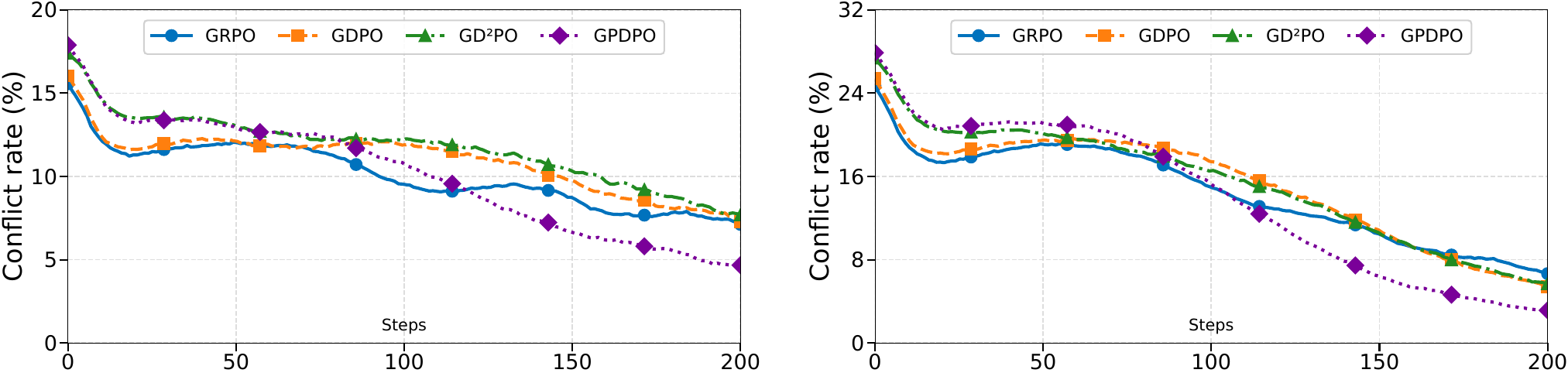}   
    \caption{
    Conflict dynamics on Qwen3-1.7B under 3R setting. \textbf{Left:} pairwise preference conflict rate over all non-tied within-group response pairs, including retained--retained pairs. \textbf{Right:} GD$^2$PO conflict rate over valid rollouts. 
    }
    \label{fig:conflict-dynamics}
\end{figure}

As shown in Tables~\ref{tab:main-2r} and~\ref{tab:main-3r}, GPDPO attains the highest Correct Acc. and Overall score across all four combinations of model scale and reward setting. Under 2R, GPDPO improves Correct Acc. over GD$^2$PO from 62.48\% to 64.99\% on Qwen3-1.7B and from 64.66\% to 66.00\% on Qwen3-4B, corresponding to gains of 2.51 and 1.34 percentage points, respectively. The improvement is particularly pronounced on the more challenging Level~3 subset for Qwen3-1.7B, where accuracy increases from 42.75\% to 48.85\%, a gain of 6.10 points. These results demonstrate that GPDPO effectively extracts useful supervision from reward conflicts, leading to improvements in the primary task objective rather than gains driven solely by auxiliary rewards or the aggregate Overall score.


The advantage persists under the more demanding 3R setting. GPDPO achieves 63.32\% Correct Acc. on Qwen3-1.7B and 65.33\% on Qwen3-4B, outperforming GD$^2$PO by 0.84 and 2.01 percentage points, respectively. Importantly, format accuracy is already near saturation across methods, yet GPDPO still improves Correct Acc. and attains the highest Overall score at both model scales. This indicates that GPDPO does more than improve auxiliary objectives such as Format Acc. and Length Rew.; by recovering pairwise supervision from filtered responses, it extracts additional learning signal that directly benefits task correctness.



\subsection{Training and Conflict Dynamics}
\label{sec:training-conflict-dynamics}

Beyond improving downstream performance, GPDPO exhibits the training dynamics: it reduces reward conflicts and achieves stronger reward optimization without accelerating entropy collapse.

\textbf{Conflict metrics.}
We measure multi-reward conflict at two levels. Following GD$^2$PO, the \emph{group-relative conflict rate} is the fraction of valid rollouts whose reward-wise advantages contain both positive and negative entries, with zeros treated as neutral. We further define the \emph{pairwise preference conflict rate} as the fraction of Pareto-incomparable pairs among all non-tied, unordered response pairs within a group. These metrics capture conflict at the aggregate rollout level and the atomic pairwise level, respectively.
\begin{table}[!t]
    \caption{Ablations on Qwen3-1.7B with 2R setting. Best Correct Acc. and Length Rew. are bold.}
    \label{tab:component-ablation}
    \centering
    \small
    \setlength{\tabcolsep}{3pt}
    \begin{tabular*}{\linewidth}{@{\extracolsep{\fill}}lrrrrr@{}}
        \toprule
        Method & \shortstack{Level 1\\Acc.} & \shortstack{Level 2\\Acc.} & \shortstack{Level 3\\Acc.} & \shortstack{Correct\\Acc.} & \shortstack{Length\\Rew.} \\
        \midrule
GD$^2$PO & 69.17 & 61.19 & 42.75 & 62.48 & 0.9793 \\
GPDPO (Strict Pareto Dominance) & 68.67 & 62.69 & 48.09 & 63.48 & 0.9650 \\
GPDPO (Scalarized Reward Ordering) & 68.42 & 49.25 & 48.09 & 61.81 & 0.9675 \\
GPDPO (Fixed Reference Model) & 71.18 & 58.21 & 39.69 & 62.81 & 0.9767 \\
GPDPO (Pareto-Guided Advantage Adjustment) & 70.68 & 61.19 & 44.27 & 63.82 & 0.9762 \\
GPDPO (Full) & 70.93 & 61.19 & 48.85 & \textbf{64.99} & \textbf{0.9812} \\
        \bottomrule
    \end{tabular*}
\end{table}

\textbf{Training dynamics.}
As shown in Figure~\ref{fig:training-dynamics}, GPDPO achieves stronger late-stage reward optimization while retaining higher policy entropy. Over steps 151--200, its mean correctness reward reaches 1.6260, compared with 1.4242 for GRPO and 1.4035 for GD$^2$PO. Its format and length rewards are also higher than GD$^2$PO (0.9566/0.9914 vs.\ 0.8902/0.9328), while maintaining higher entropy (0.0795 vs.\ 0.0746 for GRPO and 0.0719 for GD$^2$PO). Thus, the higher rewards do not coincide with faster policy collapse in this representative run.

\textbf{Conflict dynamics.}
As shown in Figure~\ref{fig:conflict-dynamics}, our method exhibits faster declines in conflict rates at both levels and achieves lower final conflict rates than GRPO, GDPO, and GD$^2$PO. This is consistent with GPDPO's design: reward-consistent Pareto relations recover auxiliary supervision from filtered responses, and each group-relative deviation can be decomposed into aggregated pairwise reward differences. The simultaneous reduction in both conflict measures suggests that recovering reward-consistent pairwise supervision is an effective strategy for alleviating multi-reward conflict.

\subsection{Ablation and Sensitivity}
\label{sec:ablation-sensitivity}


\textbf{Component analysis of GPDPO.}
To understand the contribution of each design choice in GPDPO, we conduct ablations on four variants: 
(1) {GPDPO (Strict Pareto Dominance)}, which requires strict improvement on every reward dimension instead of allowing ties;
(2) {GPDPO (Scalarized Reward Ordering)}, which replaces Pareto-based comparisons with ordering based on an aggregated scalar reward;
(3) {GPDPO (Fixed Reference Model)}, which keeps the initial policy as the reference model throughout training;
and (4) {GPDPO (Pareto-Guided Advantage Adjustment)}, which incorporates recovered Pareto relations by modifying advantages instead of directly optimizing preference margins.
All variants are evaluated on Qwen3-1.7B under 2R setting (Table~\ref{tab:component-ablation}).

\textbf{Strict Pareto dominance reduces supervision coverage.}
GPDPO with strict Pareto dominance achieves 63.48\% Correct Acc., 1.51 percentage points below full GPDPO. Strict dominance requires improvement on every reward dimension and therefore discards comparisons with ties. Standard Pareto dominance retains such pairs when one response is no worse on all rewards and strictly better on at least one, recovering additional supervision without violating reward consistency. These results suggest that allowing ties under standard Pareto dominance recovers broader yet still reward-consistent supervision, leading to more effective policy optimization.

\textbf{Scalarized ordering loses reward consistency.}
Replacing Pareto dominance with scalarized reward ordering reduces Correct Acc. from 64.99\% to 61.81\%, falling below GD$^2$PO's 62.48\%. This degradation indicates that the benefit of recovered supervision comes from preserving consistent preference directions across reward dimensions rather than from additional pairwise signals alone. 

\textbf{Reference refreshing improves optimization.}
Replacing the refreshed reference policy with a fixed reference reduces Correct Acc. to 62.81\%, compared with 64.99\% for full GPDPO. The fixed-reference variant keeps the initial policy as the reference throughout training, with \(\beta=32\) and \(\lambda=0.0003125\) selected to match the zero-margin gradient scale of full GPDPO. This result suggests that refreshing the reference policy along with the evolving sampling policy provides a more effective optimization target under the tested configuration.

\textbf{Recovered relations remain useful under alternative updates.}
GPDPO (Pareto-Guided Advantage Adjustment) improves Correct Acc. from GD$^2$PO's 62.48\% to 63.82\%, confirming that the recovered Pareto relations themselves contain useful optimization signals. Full GPDPO further improves performance to 64.99\%, suggesting that directly optimizing pairwise preference margins provides a more effective way to exploit the recovered relational supervision.

\begin{figure*}[htp]
    \centering
    \begin{minipage}[t]{0.52\linewidth}
        \vspace{0pt}
        \textbf{Robustness to Auxiliary Supervision Strength.}
        Figure~\ref{fig:lambda-sensitivity} evaluates the sensitivity of GPDPO to the parameter $\lambda$.
        Across all tested values ($\lambda=0.05$, $0.1$, $0.15$, and $0.2$), GPDPO consistently outperforms GD$^2$PO, achieving Correct Acc. scores of 63.82\%, 64.99\%, 64.66\%, and 63.15\%, respectively, compared with 62.48\% for GD$^2$PO. These results indicate that GPDPO maintains consistent improvements across a range of auxiliary supervision strengths rather than relying on a single carefully tuned coefficient.
    \end{minipage}
    \hfill
    \begin{minipage}[t]{0.43\linewidth}
        \vspace{0pt}
        \centering
        \includegraphics[
            width=\linewidth,
            height=0.53\linewidth
        ]{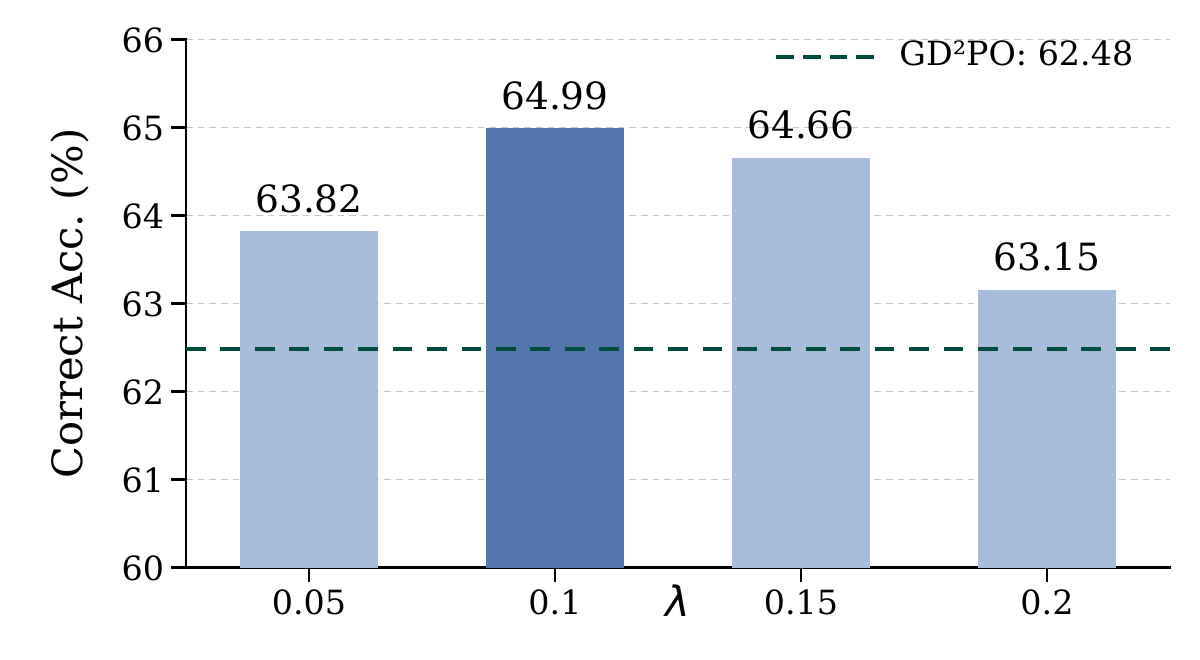}
        \vspace{-20pt}
        \captionof{figure}{
            Sensitivity of  
            $\lambda$ on Qwen3-1.7B under 2R setting.
        }
        \label{fig:lambda-sensitivity}
    \end{minipage}
\end{figure*}



\FloatBarrier

\section{Conclusion}
\label{sec:conclusion}
We presented GPDPO, a multi-reward policy optimization method that recovers reward-consistent pairwise supervision from responses discarded by conflict filtering. Our key observation is that conflicts in group-relative update signals do not necessarily eliminate reliable pairwise preferences between responses. GPDPO exploits Pareto-dominance relations within existing rollout groups to recover this supervision without additional sampling or reward evaluation. Experiments show that GPDPO consistently improves multi-reward optimization performance while reducing both group-relative and pairwise reward conflicts. These results demonstrate the value of fine-grained relational supervision for more effective multi-objective LLM alignment.


\subsection*{AI use statement}

We used generative AI tools solely to assist with manuscript writing and improve the clarity, grammar, and readability of the text. We did not use generative AI tools for any other aspect of the research. The authors reviewed and revised all AI-assisted text to ensure that it accurately reflected the methods and reported results. The authors take full responsibility for the final manuscript, including its accuracy, originality, and the integrity of its claims and conclusions.

\subsection*{Reproducibility statement}

To facilitate reproducibility, we describe the GPDPO formulation in Section 3 and provide detailed pseudocode and implementation notes in Appendix A.2. Section 4.1 introduces the datasets, base models, baselines, reward definitions, and evaluation metrics. Appendix A.1 documents the training and evaluation configurations, including hyperparameters, hardware, decoding settings, and checkpoint reporting conventions. Appendix~\ref{app:fig2-experiments} specifies the experimental setups, sampling procedures, and statistical definitions used in the conflict-filtering analysis in Figure~\ref{fig:group-mean-study}. Our implementation builds on verl and vLLM, and we will release the source code soon.

\bibliography{references}
\bibliographystyle{iclr2027_conference}

\clearpage
\appendix

\section{Implementation Details}
\label{app:implementation}
\label{app:hyperparameters}

\subsection{Detailed Settings}
\label{app:detailed-settings}

We implement GPDPO with verl and use vLLM for rollout generation.
Experiments use Qwen3-1.7B and Qwen3-4B on RLLA-4K, whose training file
contains 3,920 examples; overlength prompts are filtered during loading.
Tables~\ref{tab:training-settings} and~\ref{tab:evaluation-settings}
summarize the settings. Unless stated otherwise, all reward dimensions
have unit aggregation weights, and GPDPO uses uniform pair weights with
\(\lambda=0.1\) and \(\beta=0.1\).

\begin{table}[H]
\centering
\small
\setlength{\tabcolsep}{5pt}
\caption{Training settings for GPDPO. Batch sizes are measured in prompts.}
\label{tab:training-settings}
\begin{tabular*}{\linewidth}{@{\extracolsep{\fill}}ll@{}}
\toprule
Setting & Value \\
\midrule
Training hardware & One node, \(4\times\) NVIDIA RTX 5880 Ada (48 GB each) \\
Base models & Qwen3-1.7B, Qwen3-4B \\
Training data & RLLA-4K, 3,920 records before prompt filtering \\
Max prompt / response length & 2,048 / 1,024 tokens \\
Train batch / mini-batch size & 512 / 128 prompts \\
Rollout group size \(G\) & 4 (2,048 responses per rollout step) \\
Training steps / PPO epochs & 200 rollout steps / 1 epoch per step \\
Optimizer / learning rate & AdamW / \(10^{-6}\), constant, no warmup \\
Temperature / top-\(p\) / top-\(k\) & 1.0 / 1.0 / disabled \\
PPO clip range / dual-clip factor & \((\epsilon_{\mathrm{low}},\epsilon_{\mathrm{high}})=(0.2,0.2)\) / 3.0 \\
Policy-loss aggregation & Token mean \\
Dynamic batching & Enabled; response groups remain intact \\
vLLM GPU memory utilization & 0.60 \\
KL penalty / entropy bonus & Neither used \\
Rewards & 2R: correctness + length; 3R: additionally format \\
Pairwise loss parameters & \(\lambda=0.1\), \(\beta=0.1\); uniform pair weights \\
Reference policy & Sampling policy; refreshed every rollout step \\
Checkpoint interval / main results & Every 50 steps / selected step-200 checkpoint \\
\bottomrule
\end{tabular*}
\end{table}

Each rollout step generates \(512\times4\) responses and performs four
mini-batch updates. Correctness assigns partial credit for matching tool
names, parameter names, and values, scaled to \([-3,3]\). Length is
\(\min(\operatorname{round}(n_{\mathrm{think}}/512,2),1)\), where
\(n_{\mathrm{think}}\) counts whitespace-separated words in the
\texttt{think} block; a missing complete block receives zero. Format is
binary and checks balanced, nonrepeated, correctly ordered blocks, with
\texttt{think} always required and the other required blocks determined by GT.

\begin{table}[H]
\centering
\small
\setlength{\tabcolsep}{5pt}
\caption{Evaluation settings for the selected step-200 checkpoints.}
\label{tab:evaluation-settings}
\begin{tabular*}{\linewidth}{@{\extracolsep{\fill}}ll@{}}
\toprule
Setting & Value \\
\midrule
Evaluation hardware & One NVIDIA RTX A6000 (48 GB) \\
Dataset & Processed API-Bank, 597 questions \\
Level 1 / 2 / 3 questions & 399 / 67 / 131 \\
Decoding & Greedy, one response per question \\
Temperature / top-\(p\) & 0.0 / 1.0 \\
Max new tokens / total sequence length & 4,096 / 4,096 \\
Generation batch / tensor parallel size & 1 / 1 \\
Evaluation seed & 42 \\
Metrics & Level accuracy, Correct Acc., Length Rew.; Format Acc. for 3R \\
\bottomrule
\end{tabular*}
\end{table}

Correct Acc. is accuracy over all 597 questions. Overall is
\(\textsc{Correct Acc.}/100+\textsc{Length Rew.}\) for 2R, with
\(\textsc{Format Acc.}/100\) added for 3R. Each main-table result is a
single evaluation of a selected run. The 4,096-token total sequence limit
includes the prompt and can reduce the available generation length.

\clearpage
\subsection{Detailed Algorithm}
\label{app:algorithm}

Algorithm~\ref{alg:gpdpo} follows the three components in
Section~\ref{sec:lore}. Dominance relations and preference pairs are built
once per rollout batch. Scores and joint updates are evaluated per
mini-batch using the sampling policy fixed at the start of the rollout step.

\begin{algorithm}[H]
\caption{Group Pareto-Dominance Policy Optimization (GPDPO)}
\label{alg:gpdpo}
\small
\algrenewcommand{\algorithmicrequire}{\textbf{Input:}}
\algrenewcommand{\algorithmicensure}{\textbf{Output:}}
\algrenewcommand{\algorithmiccomment}[1]{\hspace{1em}// #1}
\begin{algorithmic}[1]
\Require Initial policy \(\pi_\theta\), prompt distribution \(\mathcal P\), reward functions \(\{r_i\}_{i=1}^{M}\),
\Statex \hspace{\algorithmicindent} group size \(G\), reward weights \(\{w_i\}_{i=1}^{M}\), loss weight \(\lambda\), margin scale \(\beta\)
\Ensure Trained policy \(\pi_\theta\)
\For{each training iteration}
  \State \(\theta_{\mathrm{old}}\leftarrow\theta,\qquad \mathcal B\leftarrow\operatorname{SampleBatch}(\mathcal P)\)
  \Statex \textbf{Stage 1: Pareto-dominance identification}
  \For{each prompt \(x\in\mathcal B\)}
    \State \(\mathcal G(x)\leftarrow\operatorname{Sample}(\pi_{\theta_{\mathrm{old}}}(\cdot\mid x),G)\)
    \State \(\ell^{\mathrm{old}}_{n,t}\leftarrow\log\pi_{\theta_{\mathrm{old}}}(y_{n,t}\mid x,y_{n,<t})\qquad(\forall n,t)\)
    \State \(\mathbf r_n\leftarrow(r_1(x,y_n),\ldots,r_M(x,y_n))\qquad(\forall n)\)
    \State \(\mu_i\leftarrow G^{-1}\sum_{n=1}^{G}r_n^i,\qquad A_n^i\leftarrow(r_n^i-\mu_i)/s_i\qquad(\forall n,i)\)
    \State \(g(y_n)\leftarrow\mathbf 1\{\operatorname{sign}(A_n^i)\operatorname{sign}(A_n^j)\geq0,\ \forall i,j\}\qquad(\forall n)\)
    \State \(\mathcal F(x)\leftarrow\{y_n\in\mathcal G(x):g(y_n)=0\},\qquad\mathcal R(x)\leftarrow\mathcal G(x)\setminus\mathcal F(x)\)
    \State \(\hat\kappa(x)\leftarrow G^{-1}\sum_{n=1}^{G}g(y_n),\qquad\widetilde A_n\leftarrow\hat\kappa(x)g(y_n)\sum_{i=1}^{M}w_iA_n^i\)
    \Statex \hspace{\dimexpr\algorithmicindent*2\relax}\Comment{Apply the GD$^2$PO conflict filter and retain its standard advantages.}
    \State \(\mathcal D(x)\leftarrow\{(y_w,y_l)\in\mathcal G(x)^2:\mathbf r_w\succ_{\mathrm P}\mathbf r_l\}\)
  \EndFor
  \Statex \textbf{Stage 2: Preference-loss construction}
  \State \(\mathcal E(x)\leftarrow\{(y_w,y_l)\in\mathcal D(x):y_w\in\mathcal F(x)\lor y_l\in\mathcal F(x)\}\quad(\forall x\in\mathcal B)\)
  \Statex \hspace{\algorithmicindent}\Comment{Select dominance edges involving at least one filtered response.}
  \State \(A\leftarrow\operatorname{MaskedWhiten}(\widetilde A,g),\qquad N_{\mathcal E}\leftarrow\sum_{x\in\mathcal B}|\mathcal E(x)|\)
  \State \(\{\mathcal M_k\}\leftarrow\operatorname{GroupPartition}(\mathcal B)\) \Comment{Keep response groups intact.}
  \For{each mini-batch \(\mathcal M_k\)}
    \State \(\ell^\theta_{n,t}\leftarrow\log\pi_\theta(y_{n,t}\mid x,y_{n,<t})\qquad(\forall n,t\text{ in }\mathcal M_k)\)
    \State \(u_\theta(y_n\mid x)\leftarrow |y_n|^{-1}\sum_{t=1}^{|y_n|}(\ell^\theta_{n,t}-\ell^{\mathrm{old}}_{n,t})\)
    \State \(\widetilde u_\theta(y\mid x)\leftarrow(1-g(y))u_\theta(y\mid x)+g(y)\operatorname{sg}[u_\theta(y\mid x)]\)
    \Statex \hspace{\dimexpr\algorithmicindent*2\relax}\Comment{Retained endpoints serve as fixed anchors in the pairwise loss.}
    \State \(\ell_{\mathrm{pair}}(y_w,y_l\mid x)\leftarrow-\log\sigma\!\left(\beta[\widetilde u_\theta(y_w\mid x)-\widetilde u_\theta(y_l\mid x)]\right)\)
    \Statex \textbf{Stage 3: Joint policy optimization}
    \State \(\mathcal L_{\mathrm{pair}}^{(k)}\leftarrow0\) \Comment{An empty pair set contributes zero.}
    \If{\(N_{\mathcal E}>0\)}
      \State \(\mathcal L_{\mathrm{pair}}^{(k)}\leftarrow N_{\mathcal E}^{-1}\sum_{x\in\mathcal M_k}\sum_{(y_w,y_l)\in\mathcal E(x)}\ell_{\mathrm{pair}}(y_w,y_l\mid x)\)
    \EndIf
    \State \(\mathcal L_{\mathrm{standard}}^{(k)}\leftarrow\operatorname{ClippedPolicyLoss}(\ell^\theta,\ell^{\mathrm{old}},A;\mathcal M_k)\)
    \State \(\mathcal L_{\mathrm{GPDPO}}^{(k)}\leftarrow\mathcal L_{\mathrm{standard}}^{(k)}+\lambda\mathcal L_{\mathrm{pair}}^{(k)}\)
    \State \(\theta\leftarrow\operatorname{OptimizerStep}(\theta,\nabla_\theta\mathcal L_{\mathrm{GPDPO}}^{(k)})\)
  \EndFor
\EndFor
\State \Return \(\pi_\theta\)
\end{algorithmic}
\end{algorithm}

\textbf{Pair construction and gradient flow.}
The implementation compares the reward-wise normalized advantages before
filtering, with tolerance \(10^{-8}\); positive per-dimension normalization
preserves Pareto ordering in exact arithmetic. Sequence scores average
log-probability ratios over valid response tokens. Retained endpoints
are detached only in the pairwise term; both endpoints receive gradients
when both are filtered. Cached sampling-policy probabilities provide the
reference without an additional fixed model.

\textbf{Loss normalization and updates.}
We implement each pair loss as \(\operatorname{softplus}(-\beta m)\),
where \(m=\widetilde u_\theta(y_w\mid x)-\widetilde u_\theta(y_l\mid x)\).
Every mini-batch and micro-batch uses the full rollout batch's pair count
as its denominator. \(\operatorname{GroupPartition}\) preserves group order;
\(\operatorname{MaskedWhiten}\) whitens retained-token advantages and leaves
filtered advantages zero. \(\operatorname{ClippedPolicyLoss}\) is GD$^2$PO's
standard objective; \(\operatorname{OptimizerStep}\) uses AdamW.
Pairs reuse the rollout responses and rewards.

\begin{table}[H]
\centering
\small
\setlength{\tabcolsep}{4pt}
\caption{Detailed results for Figure~\ref{fig:group-mean-study}.
Entries report count (percentage). The denominator is 65,536 paired label
comparisons per setting for (a,b), and 35,937 / 58,689 filtered responses
under 2R / 3R for (c,d). The four categories in (d) partition the filtered set.}
\label{tab:fig2-summary}
\begin{tabular}{llrr}
\toprule
Panel & Measurement & 2R: count (\%) & 3R: count (\%) \\
\midrule
(a) & Expansion, \(4\to64\): changed & 10,608 (16.19) & 17,756 (27.09) \\
 & Expansion, \(4\to64\): unchanged & 54,928 (83.81) & 47,780 (72.91) \\
\midrule
(b) & Peer replacement, \(G=4\): disagreement & 5,999 (9.15) & 10,434 (15.92) \\
 & Peer replacement, \(G=16\): disagreement & 6,312 (9.63) & 8,119 (12.39) \\
 & Peer replacement, \(G=64\): disagreement & 3,561 (5.43) & 3,994 (6.09) \\
\midrule
(c) & GT tool-call structure matches & 8,345 (23.22) & 16,039 (27.33) \\
\midrule
(d) & Consistently better & 4,900 (13.63) & 8,905 (15.17) \\
 & Both & 11,122 (30.95) & 16,163 (27.54) \\
 & Consistently worse & 15,752 (43.83) & 25,797 (43.96) \\
 & No consistent relation & 4,163 (11.58) & 7,824 (13.33) \\
\cmidrule(lr){2-4}
 & At least one consistent relation & 31,774 (88.42) & 50,865 (86.67) \\
\bottomrule
\end{tabular}
\end{table}
\section{Limitations}
\label{app:limitations}
Our evaluation is limited to two- and three-reward settings, as suitable benchmarks with a larger number of reward dimensions are currently unavailable. Future work will explore reward conflicts in higher-dimensional multi-reward settings and investigate scalable strategies for exploiting informative signals from conflicting responses.

\section{Additional Details for the Conflict-Filtering Analysis}
\label{app:fig2-experiments}
We supplement Figure~\ref{fig:group-mean-study} with settings and results
for Qwen3-1.7B trained with GD$^2$PO under correctness and length rewards
(2R), and with the additional format reward (3R).
\subsection{Sensitivity to the Reference Group}
\label{app:fig2-reference}

\textbf{Experimental setup.}
We use the step-50 checkpoint from each of the separate 2R and 3R training
runs. Both settings use a common set of 128 RLLA-4K training prompts.
Each model generates four target responses and a pool of 128 additional
responses per prompt, with temperature 1.0 and at most 1,024 response tokens.

\textbf{Experimental procedure.}
Target responses and rewards remain fixed. For panel~(a), we add 60 pool
responses to the original four-response group and compare the four targets'
filtering labels at \(G=4\) and \(G=64\). For panel~(b), each target is
combined with two sets of \(G-1\) peers from the pool, for
\(G\in\{4,16,64\}\). Sampling is without replacement within each set,
with overlap allowed between sets. We repeat each comparison 128 times.

\textbf{Results.}
Each rate uses \(128\times4\times128=65{,}536\) paired label comparisons
per reward setting. Expansion changes 10,608/65,536 (16.19\%) labels under
2R and 17,756/65,536 (27.09\%) under 3R. At \(G=64\), peer replacement
still produces 3,561/65,536 (5.43\%) and 3,994/65,536 (6.09\%) inconsistent
labels, respectively (Table~\ref{tab:fig2-summary}). Thus, a fixed
response's filtering decision remains sensitive to group composition.

\subsection{Information Retained in Filtered Responses}
\label{app:fig2-filtered}
\begin{table*}[t]
\centering
\small
\setlength{\tabcolsep}{5pt}
\caption{Comparison-edge statistics under 2R.
All edge counts involve at least one filtered response.
Percentages for added edges use the Pareto-edge count as the denominator.}
\label{tab:pareto-edge-analysis}
\begin{tabular*}{\linewidth}{@{\extracolsep{\fill}}lrr@{}}
\toprule
Measurement & GD$^2$PO & GPDPO (\(\lambda=0.1\)) \\
\midrule
All-dimension strict-improvement edges & 6,374 & 6,338 \\
Pareto-dominance edges & 52,421 & 52,631 \\
Edges added by allowing ties & 46,047 (87.84\%) & 46,293 (87.96\%) \\
\quad Tied correctness, improved length & 29,229 & 29,578 \\
\quad Tied length, improved correctness & 16,818 & 16,715 \\
\bottomrule
\end{tabular*}
\end{table*}
\textbf{Experimental setup.}
For panels~(c,d), we analyze rollout logs from steps 1--200 of the two
separate GD$^2$PO training runs, each with \(G=4\). Conflict filtering
identifies 35,937 responses under 2R and 58,689 under 3R.

\textbf{Experimental procedure.}
Panel~(c) measures GT matches of ordered tool-call lists, including call
order, tool names, and parameter names and values. Panel~(d) compares
each filtered response with its three original peers using Pareto dominance,
assigning it to one of the four mutually exclusive categories in
Table~\ref{tab:fig2-summary}. ``Both'' means better than some peers and
worse than others. All percentages use the full filtered set as the denominator.

\textbf{Results.}
GT matches account for 8,345/35,937 (23.22\%) and 16,039/58,689 (27.33\%)
of the 2R and 3R filtered sets, respectively. At least one consistent relation
exists for 31,774/35,937 (88.42\%) and 50,865/58,689 (86.67\%), leaving
4,163 and 7,824 responses without such relations.
These results show that filtered responses can retain task-relevant content
and consistent comparisons, supporting the recovery of pairwise supervision.

\section{Experiments on Pareto-Dominance Edges}
\label{app:pareto-edges}

We examine how allowing ties in individual reward dimensions affects the
comparison edges available for recovering pairwise supervision from filtered responses.

\textbf{Experimental setup.}
We analyze steps 1--200 of historical GD$^2$PO and GPDPO
(\(\lambda=0.1\)) training runs with Qwen3-1.7B on RLLA-4K, using correctness and length rewards (2R)
and response groups of size \(G=4\).

\textbf{Experimental procedure.}
We label responses using the training-time conflict filter and compare
two criteria: \emph{all-dimension strict improvement},
which requires higher rewards in every dimension; and \emph{Pareto dominance},
which requires no lower reward in any dimension and a higher reward in
at least one.
Both criteria use the same candidate pairs: within-group pairs involving at
least one filtered response. Retained--retained pairs are excluded, and
each edge is counted once in the better-to-worse direction.
We count edges under each criterion and identify the tied dimension in
edges admitted by Pareto dominance but excluded by strict improvement.

\textbf{Results.}
Table~\ref{tab:pareto-edge-analysis} shows that Pareto dominance retains
52,421 edges under GD$^2$PO, compared with 6,374 under all-dimension
strict improvement. The additional 46,047/52,421 (87.84\%) edges include
29,229 ties in correctness and 16,818 ties in length, with improvement
in the other reward. The GPDPO run shows a similar pattern, with
46,293/52,631 (87.96\%) Pareto edges involving such ties.

Pareto dominance allows ties in some reward dimensions while using
improvements in others to provide supervision. It retains more comparison
edges than all-dimension strict improvement, thereby providing richer
pairwise learning signals.

\end{document}